\documentclass[journal]{IEEEtran}
\usepackage{amssymb,amsmath}
\usepackage{cite}
\usepackage{graphicx}
\usepackage[caption=false]{subfig}
\usepackage[font=footnotesize]{caption}
\usepackage{psfrag}
\usepackage{url}
\usepackage[latin1]{inputenc}
\usepackage[absolute,overlay]{textpos}
\usepackage{tikz}
\usetikzlibrary{arrows,calc,decorations.markings}
\usetikzlibrary{topaths}
\usetikzlibrary{shapes,trees}
\usetikzlibrary{arrows}
\usetikzlibrary{shadows}
\usetikzlibrary{positioning}
\usetikzlibrary{matrix}
\usetikzlibrary{shapes.geometric}
\usetikzlibrary{decorations.pathmorphing}
\usepgflibrary{patterns}
\usetikzlibrary{calc}
\usetikzlibrary{fit}					
\usetikzlibrary{backgrounds}	

\usepackage{pgf}
\usetikzlibrary{arrows,automata}
\usepackage[latin1]{inputenc}

\usepackage[linesnumbered,ruled,lined]{algorithm2e}
\usepackage{amsthm}

\usepackage{tabularx}
\usepackage{makecell}
\usepackage{multirow}

\usepackage{amsmath}
\usepackage{amssymb}
\usepackage{bbm}
\usepackage{bm}
\usepackage{comment}
\usepackage[capitalize]{cleveref}

\usepackage{pgfplots}
\usepackage{float}
\usepackage{booktabs}

\pgfplotsset{compat=1.15}
\usetikzlibrary{shapes,positioning,automata, arrows.meta}

\begin{document}

\title{LoRaWAN-enabled Smart Campus: The Dataset and a People Counter Use Case}
\author{
    \IEEEauthorblockN{Eslam Eldeeb and Hirley Alves}
    \thanks{The authors are with Centre for Wireless Communications (CWC), University of Oulu, Finland. Email: firstname.lastname@oulu.fi.} 
    \thanks{This work is partially supported by Academy of Finland 6Genesis Flagship (Grant no. 346208) and FIREMAN (Grant no. 326301).}
}

\maketitle

\begin{abstract}
IoT has a significant role in the smart campus. This paper presents a detailed description of the Smart Campus dataset based on LoRaWAN. LoRaWAN is an emerging technology that enables serving hundreds of IoT devices. First, we describe the LoRa network that connects the devices to the server. Afterward, we analyze the missing transmissions and propose a k-nearest neighbor solution to handle the missing values. Then, we predict future readings using a long short-term memory (LSTM). Finally, as one example application, we build a deep neural network to predict the number of people inside a room based on the selected sensor's readings. Our results show that our model achieves an accuracy of $95 \: \%$ in predicting the number of people. Moreover, the dataset is openly available and described in detail, which is opportunity for exploration of other features and applications. 
\end{abstract}

\section{Introduction}\label{sec:introduction}
The Internet-of-Things (IoT) demands serving a massive number of limited-power devices with high reliability and low costs~\cite{9845353}. In addition, the IoT connectivity landscape is quite scattered, comprising many low-power wide-area (LPWA) technologies, which can be split into two large groups, i.e., cellular (e.g., NB-IoT, LTE-M, 5G-RedCap) and non-cellular technologies (e.g., Long-Range Wide Area Network (LoRaWAN) and SigFox). In this context, however, cellular networks suffer from multiple limitations in terms of cost and energy consumption~\cite{7389044}, which limits the applicability in extremely low-power and/or low-cost scenarios. Thus, LoRaWAN and SigFox~\cite{7815384} have been considered suitable candidates in such cases due to their long range, low power demands, and low costs~\cite{8480255}.

LoRa is wide area network wireless technology developed by Cycleo and patented by Semtech~\cite{LoRa_Patent}. It is low power, low range, low data rate-based technology that operates in industrial, scientific, and medical (ISM)
bands over the frequency bands 863 to 870 MHz in the EU, and uses Chirp Spread Spectrum (CSS) as its modulation scheme~\cite{8617880}. The duty cycle ranges between $0.1 \%$ to $1 \%$ based on the sub-band. The LoRa network usually consists of end-LoRa devices, a gateway, and a network server, where the network uses star of stars topology~\cite{8883217}.

LoRaWAN is an emerging LPWA technology that uses LoRA as its PHY layer~\cite{7803607} due to its high range and low power demands. It shows huge scalability supporting hundreds to thousands of devices within the network~\cite{8430542}. Smart campus is defined in~\cite{8267975} as exploiting IoT service providers to enable services over the internet. It relies on collecting the readings from sensors and connected devices to provide a comfortable atmosphere and enhance the experience of the students and teachers on the campus. LoRa technology can be used in smart campuses to serve and collect data from different sensors~\cite{8288154}. These sensors can measure various parameters, such as CO2, temperature, and humidity. The LoRa technology is used to carry the sensor readings to the server through a gateway. These readings are then processed on the server end.

\begin{figure}[t!]
    \centering    
    \includegraphics[width=1\columnwidth]{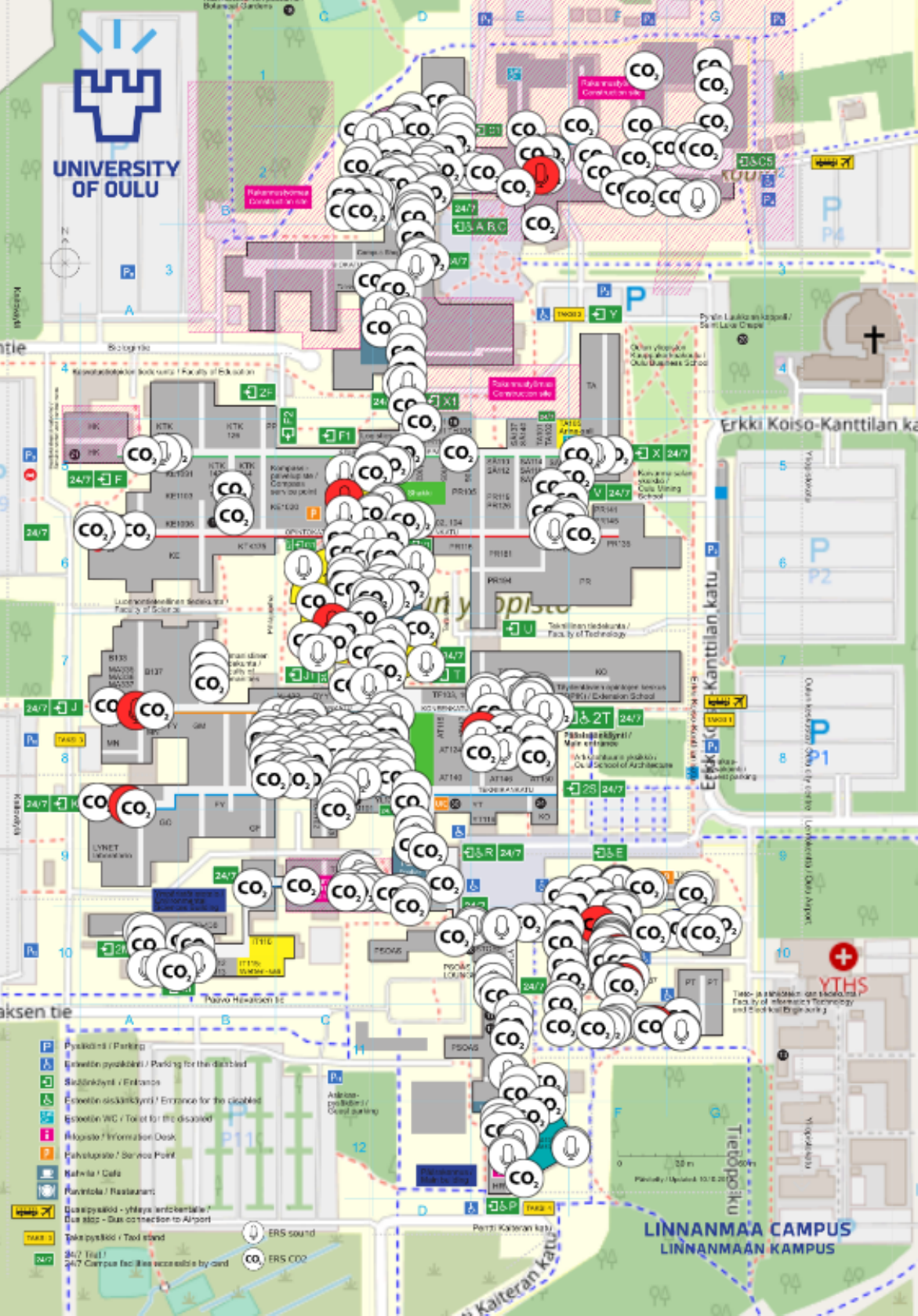} \vspace{1mm}
	\caption{The smart campus IoT sensor network.} \vspace{0mm}
	\label{MAP}
\end{figure}

In this work, we present the Smart Campus dataset~\cite{dataset_upload}. The Smart Campus dataset is an open research dataset focusing on industry-academia collaboration, network building, research-based campus development, and piloting novel Smart Campus services~\cite{5GTN_Dataset_citation}. It consists of hundreds of sensors deployed on the campus of the University of Oulu, as shown in Fig.~\ref{MAP}. It uses LoRa technology to carry the sensor readings to the server. The smart campus dataset can be used in many applications, such as time-series forecasting, anomaly detection, spatial correlation, and occupancy estimation of certain spaces. In addition, various data analysis techniques can be applied to the LoRa parameters, such as battery consumption, power analysis, and failure transmission analysis.

LoRa can suffer packet losses in massive deployments due to transmission outages or collisions~\cite{8081678}. This leads to the loss of important sensor readings that further affect the data analysis. According to \cite{4292217}, the missing values have three main drawbacks: \textit{(i)} low efficiency, \textit{(ii)} data analysis is more difficult and complex, and \textit{(iii)} bias the results towards the existing data over the missing one. Therefore, handling the missing values in the dataset is required to guarantee a robust data analysis. There are many approaches to handle the missing data in the dataset, such as dropping its value, using linear or polynomial interpolation, and even machine learning tools to fill the missing gaps~\cite{9608954}.

During the pandemic, new types of needed restrictions arise~\cite{HAN20201525}. The number of people in closed rooms should be controlled to limit the spread of the deadly virus~\cite{pokhrel2021literature}. However, there might be some violations in gathering indoors, where the number of people may exceed the allowed limit. Therefore, there should be continuous monitoring of the number of people inside the closed rooms. Nowadays, smart air ventilation systems exist in every smart residual building~\cite{guyot2018smart}. Monitoring the number of people indoors is important to guarantee effective smart air ventilation systems. In this work, we use the collected readings of the sensors to predict the number of people inside a room after handling the missing values in the dataset. We note that this use case is timely and illustrative of the potential exploitation of this rich dataset.

\subsection{Related Work} \label{sec:related_work}
Herein, we summarize the existing literature discussing the LoRa technology and the work done on similar networks. To begin with, the authors in~\cite{s18113995} present a full overview of the LoRa technology from the standardization, physical layer, and network layer perspective, whereas~\cite{8550722} presents a comparison between the LoRa other LPWAN technologies. In~\cite{8019271}, the authors discuss the LoRa as the emerging technology in the massive IoT network by presenting a health-care practical use case. The work in \cite{8372906} discusses the effect of message space and time diversity on the success probability using message replication and multiple antennas. The authors in \cite{8959186} compare the replication, coded, and hybrid transmission. The hybrid transmission has the lowest outage probability. A LoRaWAN simulator is presented in~\cite{9083800} to study the sustainability performance of LoRa networks in terms of coverage and throughput.

Counting the number of people has been discussed widely in the literature. In $1994$, Gary developed a real-time people counter using a fixed-camera~\cite{Conrad1994ARP}. The authors in~\cite{10.1145/2733373.2806337} developed a deep convolution neural network (CNN) to count the number of people in extremely crowded areas for video surveillance. Away from the computer vision-based people counter, the work in~\cite{635472} discusses using infrared (IR) sensors to count the number of people crossing a door. Moreover, the authors in~\cite{8293759} succeed in counting accurately up to $5$ people in a room using the Wi-Fi signals, whereas the authors in~\cite{7155631} suggest counting the number of people based on Wi-Fi signals. Their model shows accurate results up to $93 \: \%$ counting accuracy indoors.

\subsection{Contributions}
This work presents a detailed analysis of the Smart Campus dataset. Our main contributions are:
\begin{itemize}
    \item We present a description and a preliminary analysis of the Smart Campus dataset.

    \item We address the required pre-processing and data imputation steps before the data analysis. We identify transmission failures of the devices and how to handle such missing data using different approaches, such as interpolation and k-nearest neighbor (KNN).

    \item In addition, we build a long short-term memory (LSTM) model to be used as a time-series predictor to forecast the future readings of the sensor as an evaluator of the failure handling techniques.

    \item Moreover, we formulate a deep neural network to predict the number of people inside a room using the readings of the devices. 
    
    \item Simulation results show the high accuracy of predicting the number of people, indicating any exceeding of the limitations in closed rooms, particularly relevant during the pandemic.
\end{itemize}

Owing to reproducibility and openness, the dataset is openly available at \cite{5GTN_Dataset_citation}, and all analysis in this work can be accessed through the GitHub repository~\cite{People_Notebooks}.

\subsection{Outline}
The rest of the paper is organized as follows: Section~\ref{sysmodel} illustrates the dataset and system model details. Section~\ref{People_Counter_Section} presents the proposed data imputation and data analysis. Section~\ref{KPIs} illustrates the evaluation metrics. Section~\ref{results} depicts the simulation results, and Section~\ref{conclusions} concludes the paper.

\section{Dataset Description and Preliminary Analysis}\label{sysmodel}
\begin{figure}[t]
    \centering
    \subfloat[The LoRa Parameters dataset \label{LoRA_Dataset}]{\includegraphics[width=0.45\textwidth]{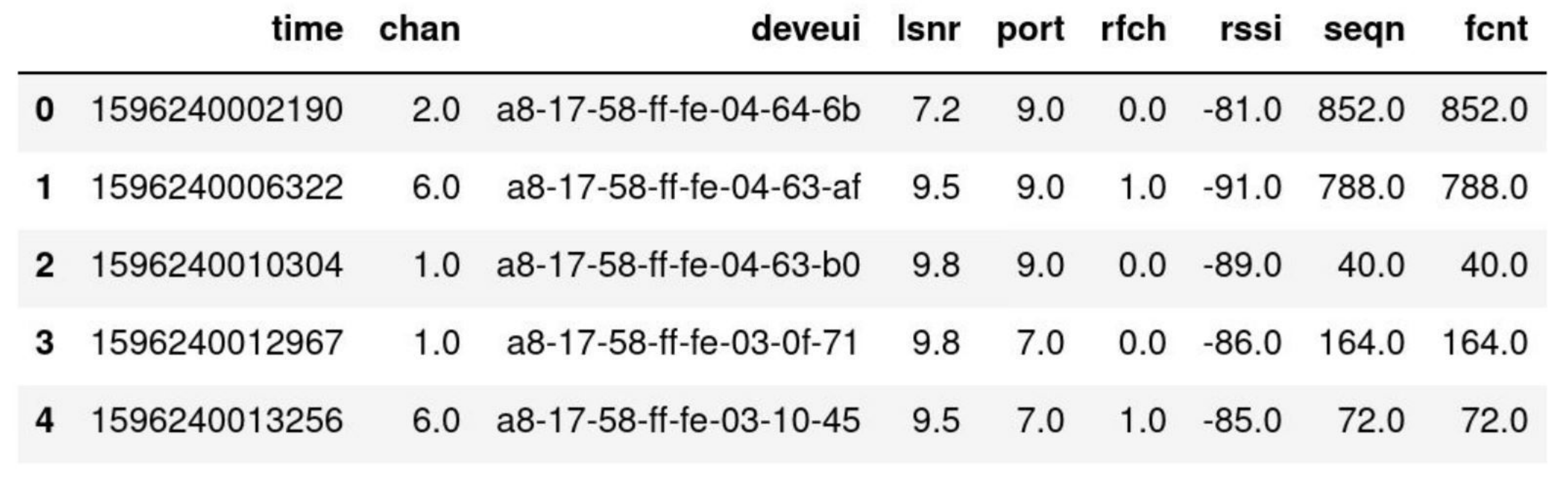}}
    \hskip -1.9ex
    \subfloat[The sensor readings dataset \label{Sensors_Dataset}]{\includegraphics[width=0.45\textwidth,trim={0 0 0 0},clip]{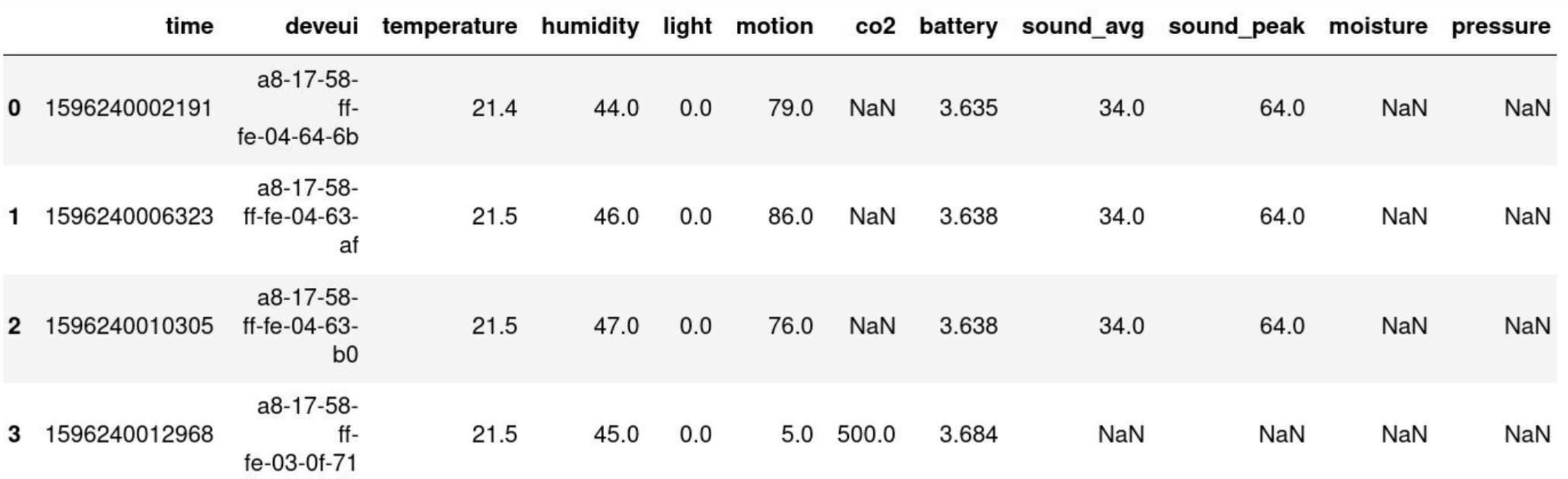}}
    \hskip -1.9ex
    \caption{A snapshot from the LoRa parameters dataset and the sensors readings dataset.}
    \label{DATASETS} \vspace{-0mm}
\end{figure}

The Smart Campus dataset summarizes the smart campus IoT sensor network measurements that decompose hundreds of low-power sensors scattered across the University of Oulu campus (indoor) and the botanical garden (outdoor). The smart campus IoT sensor network consists of $462$ devices scattered across $135,000$ $\text{m}^2$ area. It consists of two datasets, namely, \textit{(a) LoRa parameters dataset} that presents the physical layer characteristics of the LoRa network, and \textit{(b) Sensors readings dataset} that presents the measurements of the sensors. The former, as shown in Fig.~\ref{LoRA_Dataset}, consists of the time stamp of transmission, the channel used in transmission (there are $7$ available channels), the device extended unique identifier DevEUI), the LoRa signal-to-noise ratio (LSNR) of the transmission, the port that is used to distinguish between messages, the binary RF chain value (RFCH), the received signal strength indicator (RSSI), and the frame counter (FCNT). Table~\ref{LoRA_Range} presents the range of values of each parameter in the LoRa parameters dataset. In addition, As shown in Fig.~\ref{Sensors_Dataset}, the latter has the physical measured quantities besides the time stamp and the DevEUI. The sensors are divided into three types: \textit{i) CO2 devices} measure CO2 levels, motion and light, \textit{ii) sound devices} measure sound average, sound peak, motion, and light, and \textit{iii) moisture devices} measure pressure and moisture. In addition, all devices measure temperature and humidity and monitor their battery levels. The physical quantities monitored by a device have a $nan$ reading. Among the $462$ devices, there are $326$ CO2 devices, $119$ sound devices, and $17$ moisture devices.

\begin{table}[!t]
\centering
    \caption{The LoRa parameters dataset range of values.}
	\label{LoRA_Range}
\begin{tabular}{|l|l|}
\hline
\textbf{Parameter} & \textbf{Range} \\\hline

Channel & $\left[0 \: \: 6\right]$ \\\hline

LSNR & $\left[-22.5 \: \: 10\right] \text{dB}$\\\hline

Port & $\left[1 \: \: 12\right]$ \\\hline

RFCH & $\left[0 \: \: 1\right]$ \\\hline

RSSI & $\left[-53 \: \: -120\right] \text{dB}$ \\\hline

FCNT & $\left[0 \: \: 65535\right]$ \\\hline

\end{tabular} \vspace{-0mm}
\end{table}

\begin{figure}[t!]
    \centering    \includegraphics[width=0.8\columnwidth]{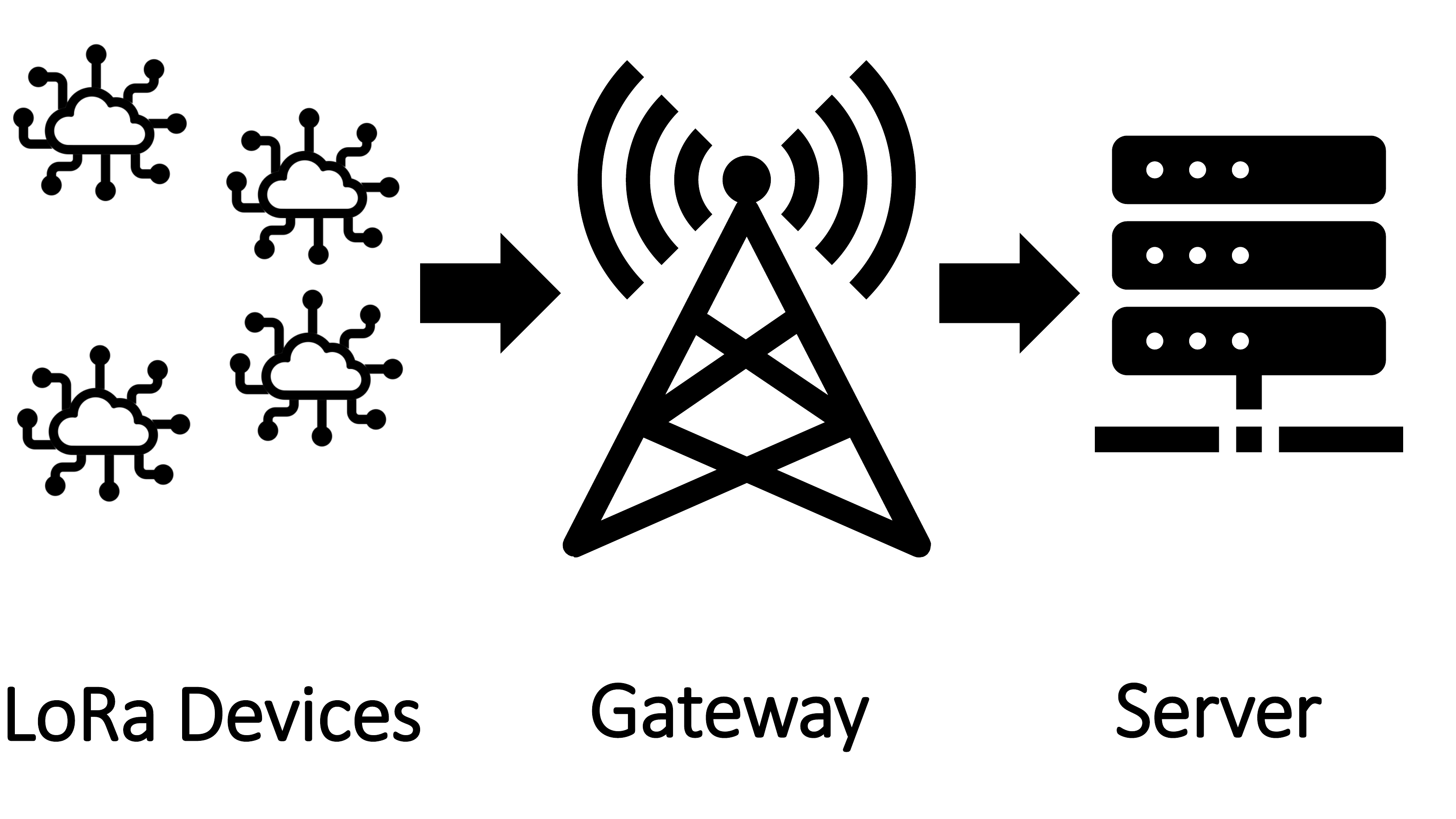} \vspace{1mm}
	\caption{The system model: The LoRa devices transmit their packets to a gateway, which relays the collected data to the server.} \vspace{0mm}
	\label{System_Model}
\end{figure}


The devices save the date and the time of their readings. The date format is $yyyy.mm.dd$, where $yyyy$ represents the year in four digits, $mm$ is the month in two digits, and $dd$ represents the day in two digits. The time format is $hh.mm.ss.msmsms$, where $hh$ represents the hour in two digits, $mm$ is the minute in two digits, $ss$ is the second in two digits, and $ms$ is the millisecond in three digits. We represent the set of devices as $\mathcal{D}=\{1,2,\cdots,D\}$, where the coordinates of device $d$ is $c_d = (x_d,y_d)$ and its height is $h_d$. Every sensor should save its reading every $15$ minute and transmit it to a gateway $G$ with the coordinates $c_G = (x_G,y_G)$. The gateway relays the received readings to the dedicated server, as shown in Fig.~\ref{System_Model}. A LoRa network connects the devices, the gateway, and the server.



\begin{figure*}[t]
    \centering
    \subfloat[An example CO2 device\label{CO2_example}]{\includegraphics[width=0.33\textwidth]{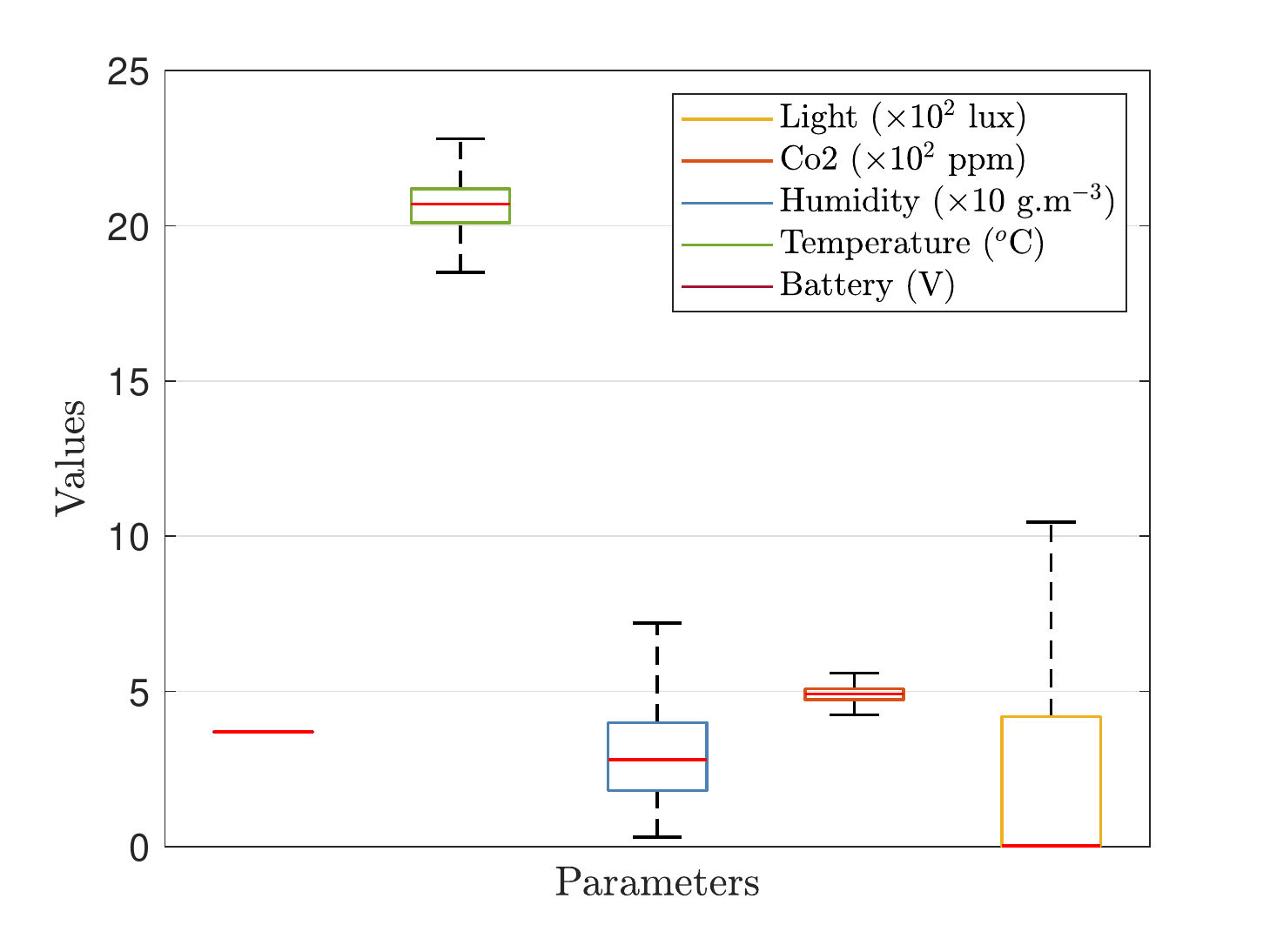}}
    \hskip -1.9ex
    \subfloat[An example sound device \label{Sound_example}]{\includegraphics[width=0.33\textwidth,trim={0 0 0 0},clip]{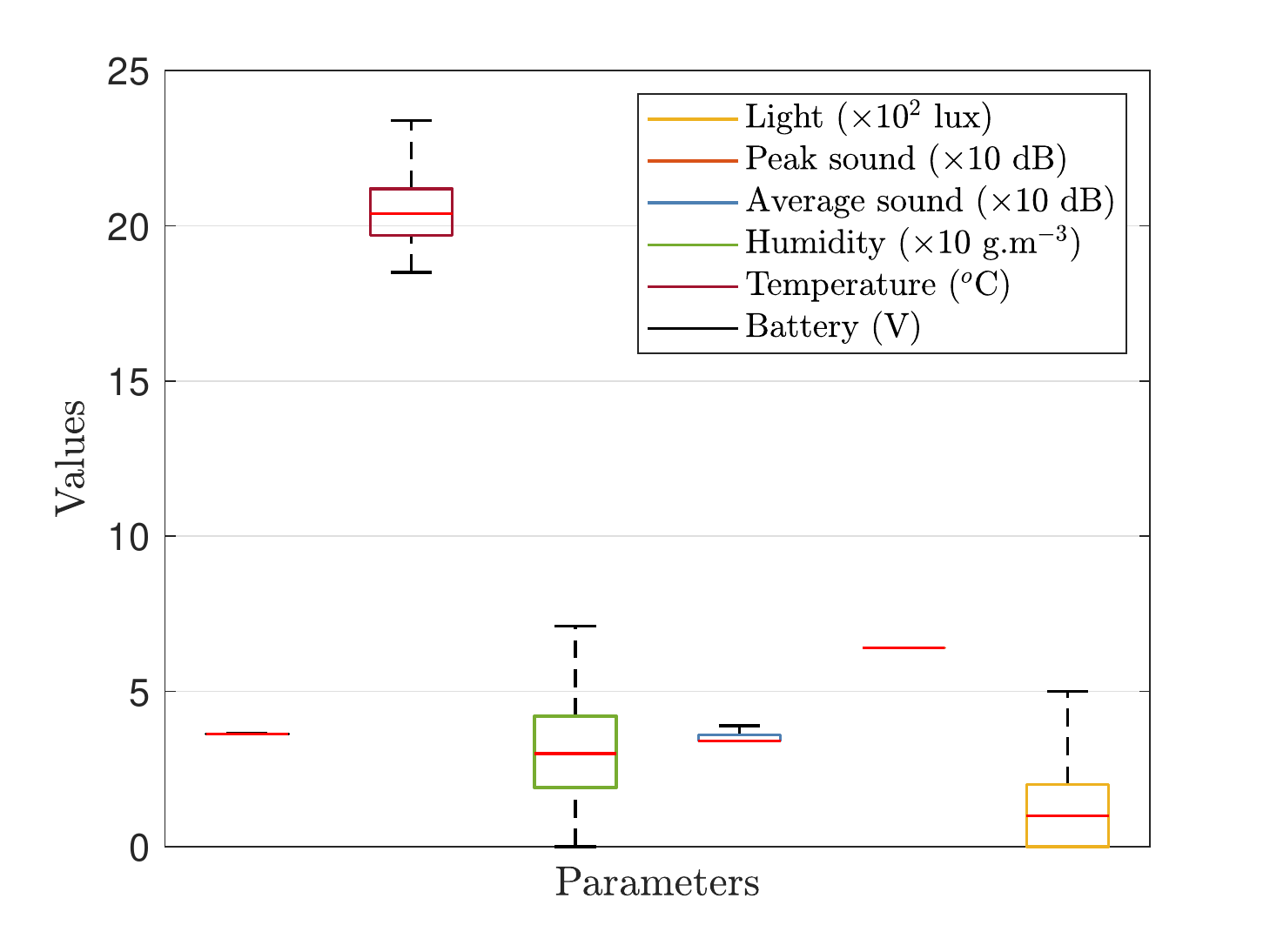}}
    \hskip -1.9ex
    \subfloat[An example moisture device \label{Moisture_example}]{\includegraphics[width=0.33\textwidth,trim={0 0 0 0},clip]{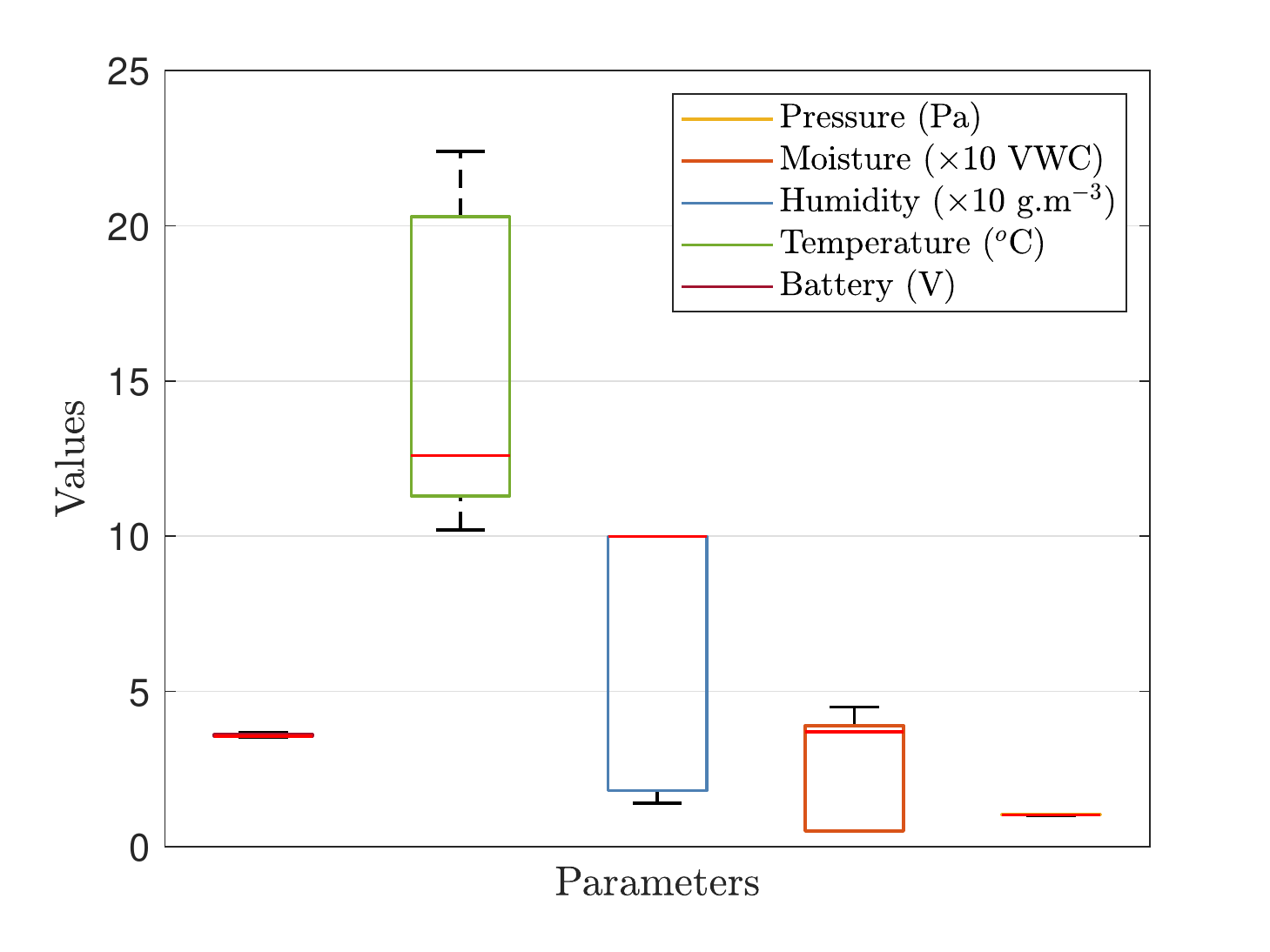}}
    \hskip -1.9ex
    \caption{The parameters values of an example CO2 device, sound device, and moisture device.}
    \label{Example} \vspace{-0mm}
\end{figure*}

\begin{figure*}[t]
    \centering
    \subfloat[Max \label{max}]{\includegraphics[width=0.33\textwidth]{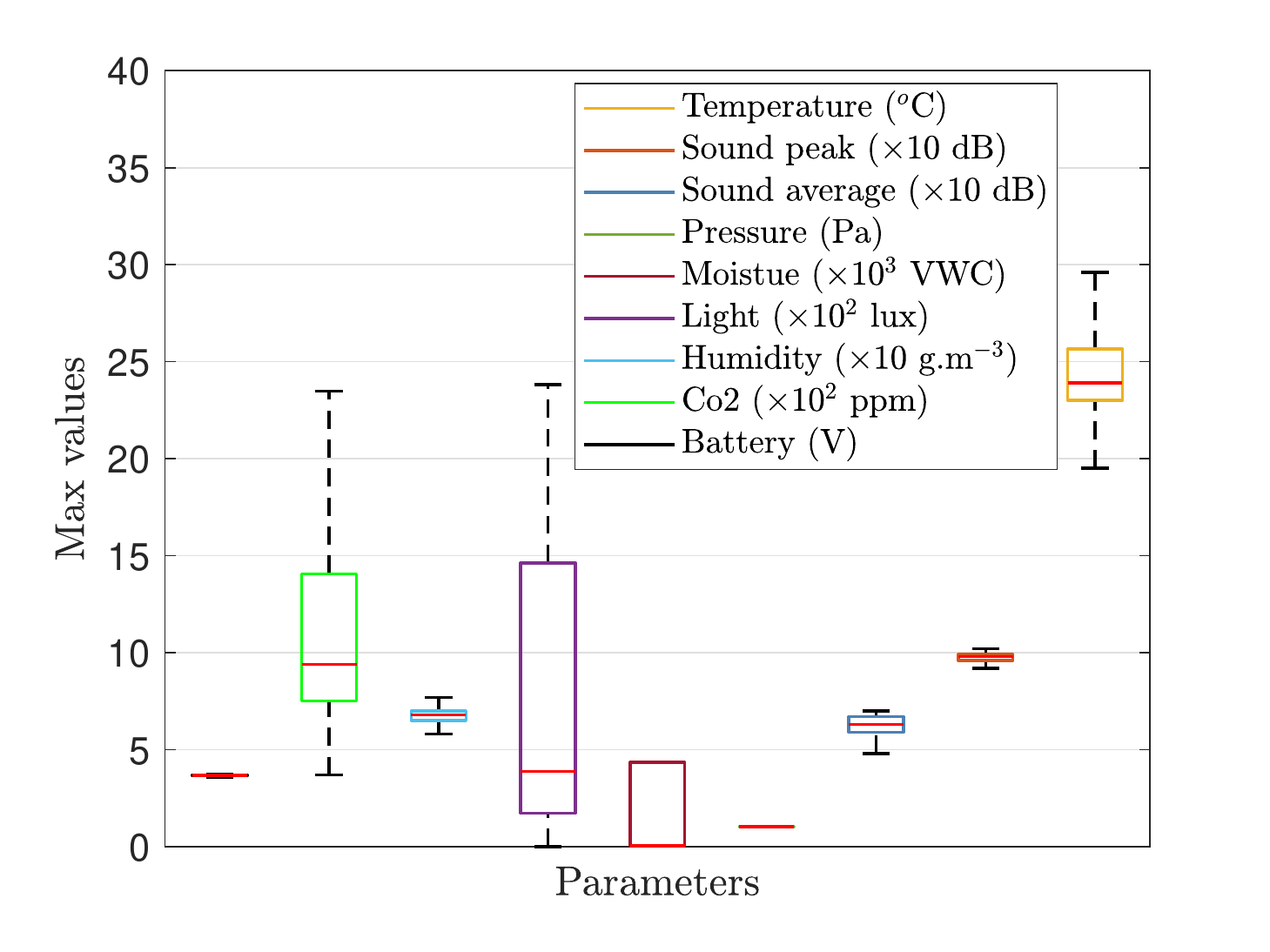}}
    \hskip -1.9ex
    \subfloat[Min \label{min}]{\includegraphics[width=0.33\textwidth,trim={0 0 0 0},clip]{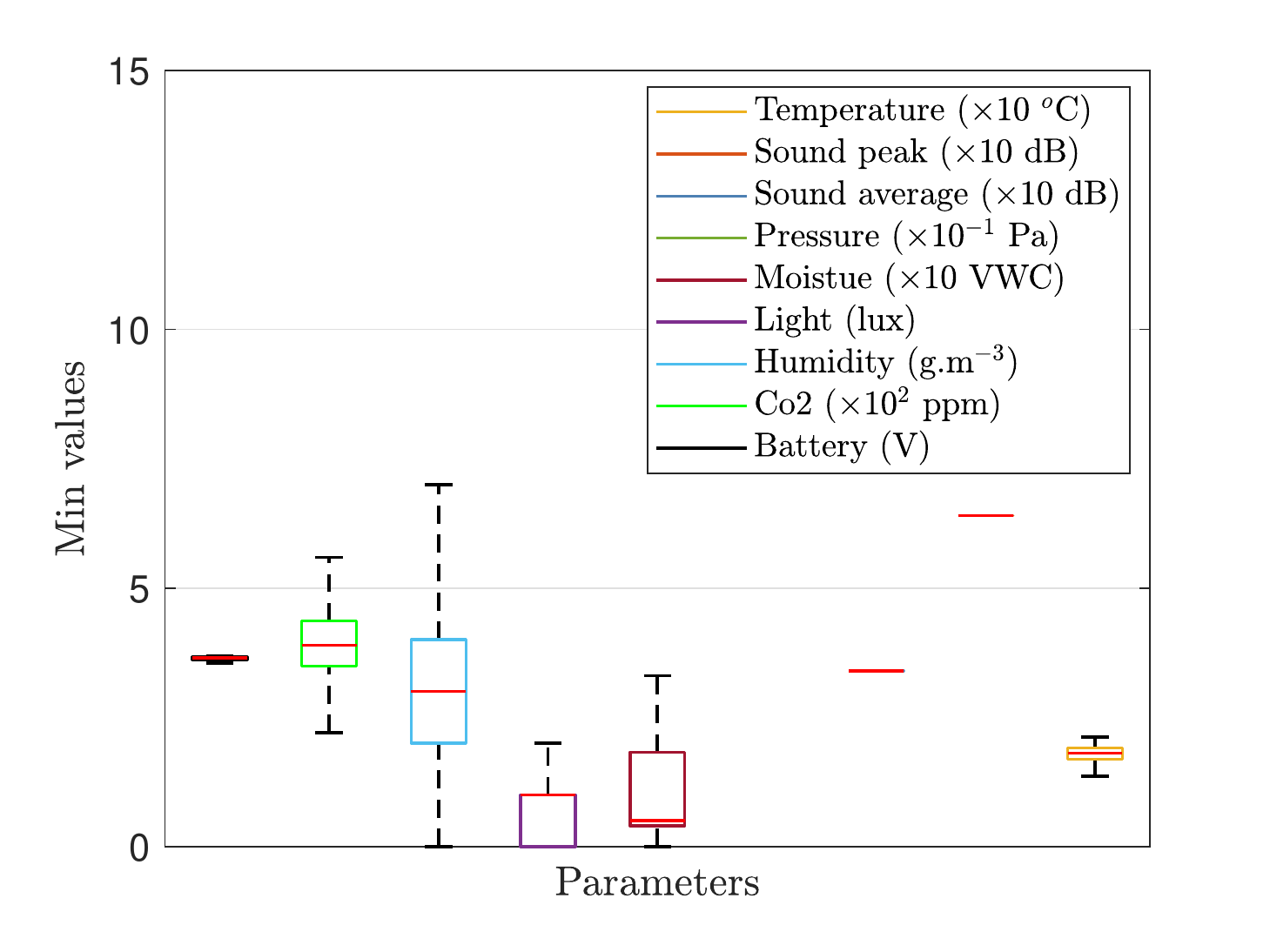}}
    \hskip -1.9ex
    \subfloat[Mean \label{mean}]{\includegraphics[width=0.33\textwidth,trim={0 0 0 0},clip]{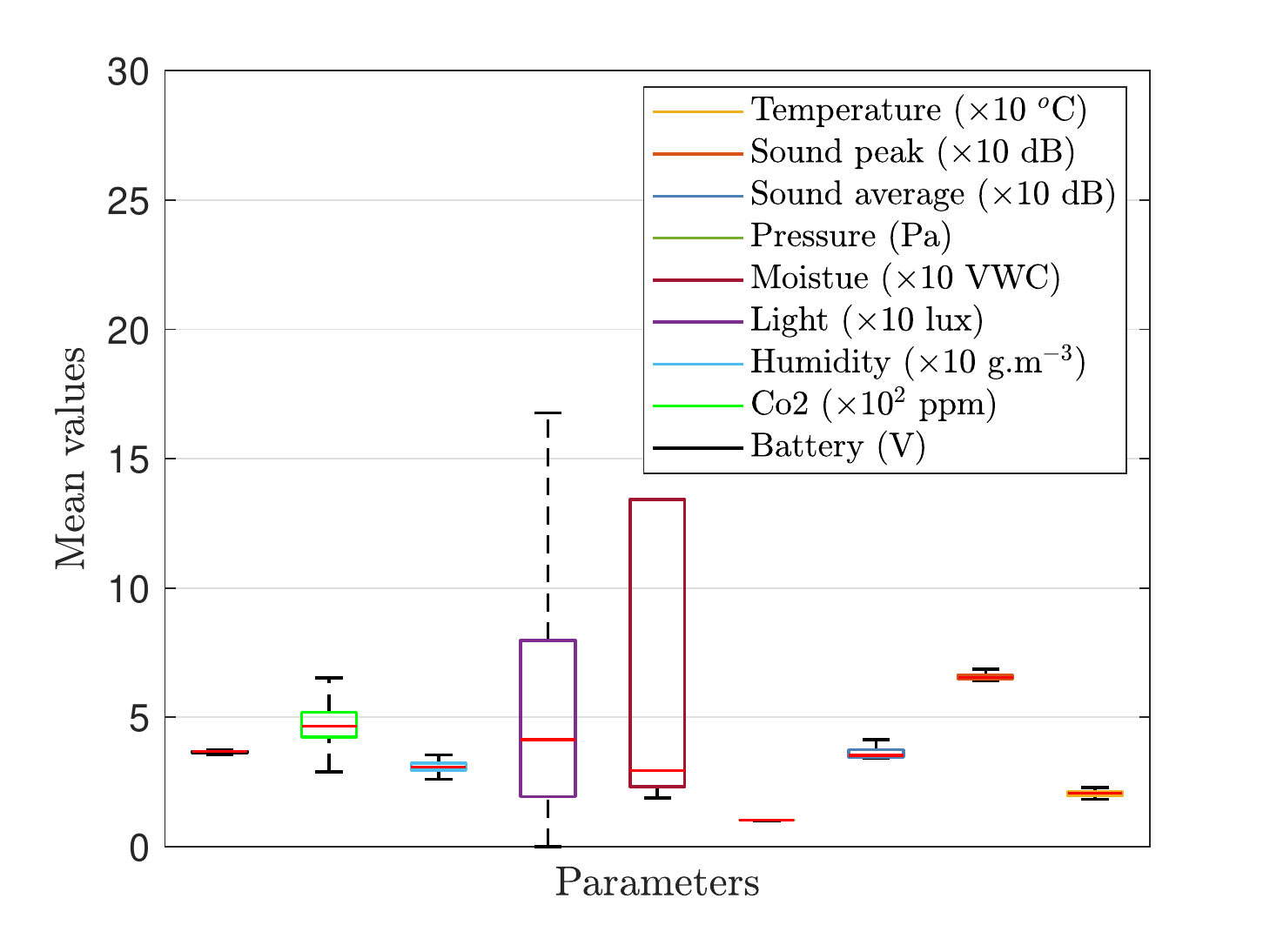}}
    \hskip -1.9ex
    \caption{The max, min, and mean analysis of the parameter values of all the devices.}
    \label{CO2_Anaysis} \vspace{-0mm}
\end{figure*}

Fig.~\ref{Example} shows the parameter readings of some example devices. First, we remove all the outliers of these readings and then present them in a box-plot view. The outliers are values with a sudden drop or sudden increase over the normal readings of the device and the sane physical values of a certain parameter. These outliers have many sources, such as high channel noise, an outage in the transmission, and unknown sudden internal noise. An important pre-processing step is to handle such outliers as it affects the post-processing analysis. Fig.~\ref{CO2_example} depicts the range of light readings, CO2, humidity, temperature, and battery consumption for an example Co2 device inside a closed room. For instance, the temperature values varies from $15 \: ^o C$ and $39 \: ^o C$, whereas the CO2 values has a range of $424$ ppm and $600$ ppm. The light varies from $0$ lux to $1060$ lux. The changes in the CO2 and light values can be used as indicators of someone inside the room and the number of people inside the room at a time. Fig.~\ref{Sound_example} describes the range of light readings, peak sound, average sound, humidity, temperature, and battery consumption for an example sound device inside a closed room. We can notice that the average and peak sounds slightly change their values as they vary around $45$ dB and $70$, respectively.

Moreover, Fig.~\ref{Moisture_example} presents the range of readings of pressure, moisture, humidity, temperature, and battery consumption for an example moisture device in the soil of the garden. The pressure readings have a very slight change in the soil, whereas the moisture ranges from $5$ VWC to $45$ VWC. The battery consumption readings seem to change slightly in the three examples.

Furthermore, we present a precise statistical analysis of each device's maximum, minimum, and mean values in Fig.~\ref{CO2_Anaysis}. The behavior of the CO2 readings differs in terms of each device's max, min, and mean values. This is because the sensors behave differently indoors and outdoors, and each indoor room has a different size, different number of people inside, and different ventilation systems. This is also applied to the light measurements as each room has a different light system, which can be more intense in one than the other, and the soil moisture, which is highly affected by the rain and the weather. The max sound records differ as the number of people, presentation type, and other parameters affect the max values of both average and peak sound. In contrast, the minimum and mean values are almost the same, and the sensors capture almost the same values when there is no sound around them.

\section{Illustrative application: proposed people counter model}\label{People_Counter_Section}
\begin{figure}[t!]
    \centering    \includegraphics[width=1\columnwidth]{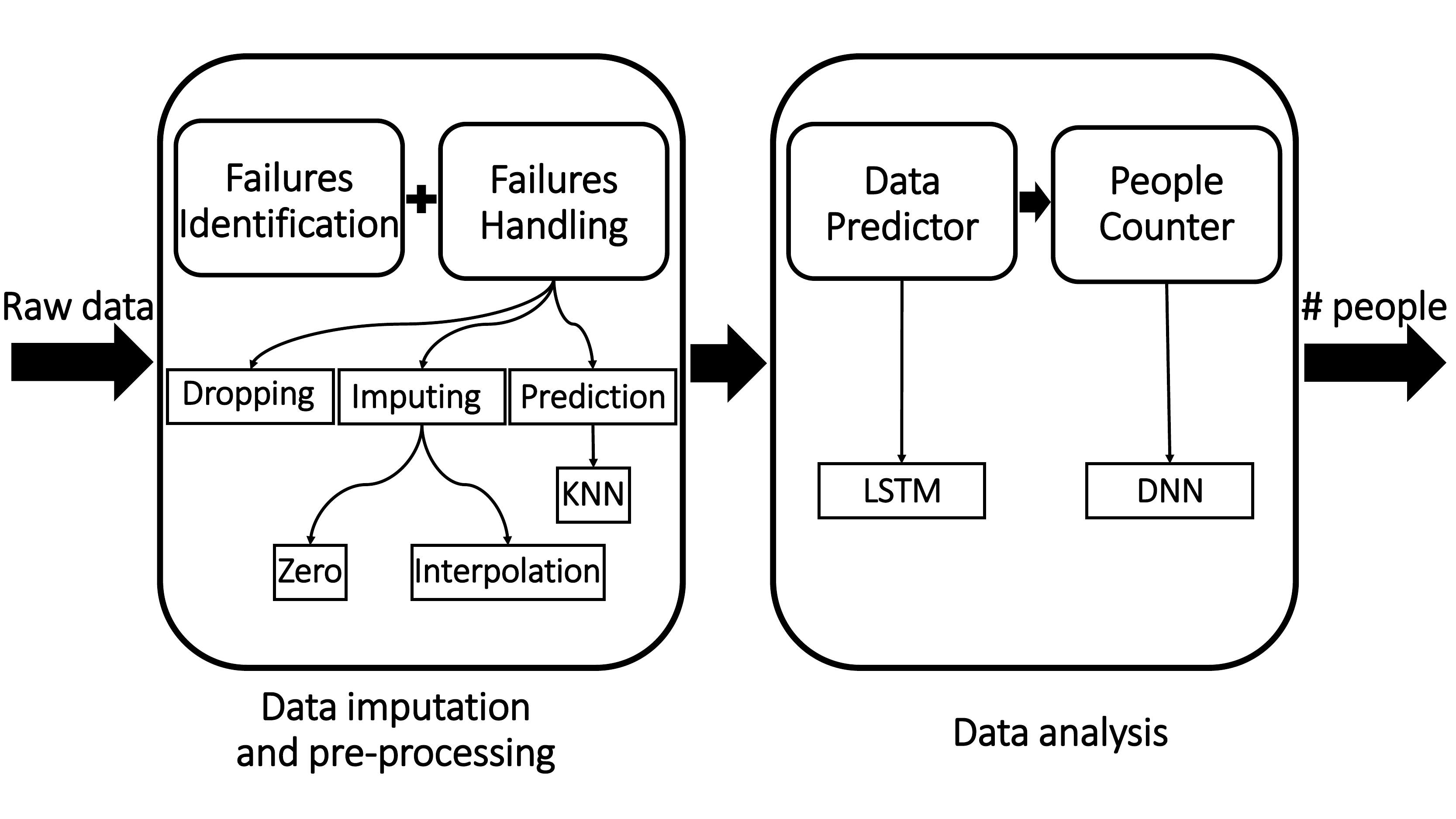} \vspace{1mm}
	\caption{The proposed people counter model. First, the data imputation and pre-processing phase involve failure identification and handling. Then, the data analysis phase consists of the data predictor, and people counter.} \vspace{0mm}
	\label{proposed_Model}
\end{figure}

In this section, we introduce the data imputation techniques performed on the described dataset, followed by the data analysis required to predict the number of people inside a room. Herein, we propose a people counter model. We first pre-process the sensor readings by detecting missing data in the dataset due to transmission failures and filling in the missing data. Afterward, we predict the future readings of the sensors and formulate the people counter using the output of the data predictor. We focus on CO2 readings to validate the presented scheme. Fig.~\ref{proposed_Model} illustrates the proposed people counter model.

\subsection{Data Imputation and Pre-Processing}\label{DataImput}

In the beginning, we detect the outliers and remove them. In addition, we identify the missing transmission (failures), and finally, we introduce several solutions to handle the missing transmissions and fill the gaps. After choosing the most effective technique to handle the missing values, we normalize and prepare the data for the data analysis phase.

\subsubsection{Failures Identification}\label{subsec:FailIdent}
A failure $f$ occurs if an outage exists in the transmission between the device and the gateway or between the gateway and the server, causing packet loss. Each LoRa device should report its readings every $15$ minute to the gateway, which relays the readings to the server. A failure is identified if there is a missing packet $15$ minutes before the last transmission. The FCNT parameter in the LoRa parameters dataset is also used to identify the failures. A transmission failure occurs whenever there is a missing packet count in the FCNT parameter.

\subsubsection{Failures Handling}\label{subsec:FailHand}
Before performing data analysis, we must fill in the missing data in the datasets resulting from the failure transmissions. There are various methods to handle the missing data from the datasets:
\begin{enumerate}
    \item Dropping: remove the reading and assume the non-existence of that particular missed transmission.
    
    \item Imputing: try to manipulate particular values in the missing places. The simplest approach is to replace the missing values with zero values. This approach is simple but not efficient as it affects the accuracy of any analysis performed over the dataset. Linear interpolation is a method to fit the known data points with a linear curve to find unknown intermediate data points 
    \begin{equation}
        y = a_0 + a_1 x,
    \end{equation}
    where $a_0$ and $a_1$ are the intercepts. It is a simple and efficient approach, especially with datasets with few missing values. Polynomial interpolation generates a polynomial function that fits the data points
    \begin{equation}
        y = a_0 + a_1 x + a_2 x^2 + ... + a_n x^n,
    \end{equation}
    where $n$ is the order of the polynomial function.

    \item Prediction: use the known data points to predict the unknown missing values. The KNN is a proximal interpolation that predicts the missing values by finding the k-most similar data points. To find the best value for $k$, we collect a group of data with no missing values and force some known data to be unknown. Afterward, we apply KNN with different values of $k$ to predict the missing values; then, we calculate the mean square error (MSE) between the actual data and the predicted one. The optimized value $k^*$ is the one that has the lowest MSE.
\end{enumerate}

\subsection{Data Analysis}\label{sec:DataAnal}
In this section, we present the proposed data analysis techniques. First, we evaluate the proposed techniques to handle the missing values by predicting a batch of clean data of the sensor's readings in the future and comparing it to the true readings. Then, we utilize the sensor readings to predict the number of people inside a room. 

\subsubsection{Data Predictor}\label{subsec:DataPred}
In this section, we propose a method to predict the future readings of the devices based on the known present data. We first fit the failure handling methodologies proposed in the previous subsection and then predict the future readings. This problem is considered a time-series forecasting problem. LSTM is a well-known architecture for solving time-series problems~\cite{article_LSTM}. It overcomes the vanishing gradient problems of the basic recurrent neural networks architecture, and thus, LSTM is very efficient in solving long sequences time-series forecasting problems~\cite{279181}. At each time $t$, it receives a sequence of data and returns two outputs: the short-term memory $h_t$ and the long-term memory $C_t$. The LSTM consists of 4-gates: \textit{i) forget gate,} which ignores the irrelevant present data, \textit{ii) learn gate,} which learns the relation between the data points in a given sequence, \textit{iii) remember gate,} which utilizes the forget gate and the learn gate outputs to update the long-term memory, and \textit{iv) use gate,} which updates the short-term memory~\cite{9504554}. The LSTM update equations are formulated as follows
\begin{align}\label{LSTM_Vec1}
\begin{pmatrix} i_t \\ f_t \\ o_t \\ C_{t_i} \end{pmatrix} &= \begin{pmatrix} \sigma \\ \sigma \\ \sigma \\ tanh \end{pmatrix} \: W \: \begin{pmatrix} h_t \\ h_{t-1} \end{pmatrix}, \\
C_t &= f_t \: \odot \: C_{t-1} + i_t \: \odot \: C_{t_i}, \\
h_t &= o_t \: \odot \: tanh(C_t),
\end{align}
where $i_t$, $f_t$, and $o_t$ are the outputs of the learn gate, forget gate, and use the gate, respectively. In addition, $C_{t_i}$ is the initial long-term memory, $\sigma$ is a sigmoid function, $W$ is the weighs vector, and $\odot$ is a point-wise multiplication.

\subsubsection{People Counter}\label{subsec:PeopleCnt}
We rely on the readings of the devices to predict the number of people in a closed room. We track and save the readings of a device in a room (the features) for a certain period and the number of people inside the room (the labels) during the same period. We chose a meeting room in the university that should be reserved on campus. To reserve the room, the person should estimate the number of people in the meeting. We add the number of people in the dataset corresponding to the sensor readings inside that particular room at the time of the reservation. We divide the collected data into training data and testing data. We try to build a model using the training data to efficiently predict the number of people in the testing data and then be generalized for any data for the same device in the same room. This problem is considered a classification problem. Neural networks are one of the most powerful tools used in classification problems. 

\begin{figure*}[t]
    \centering
    \subfloat[LSTM training loss \label{Models_loss}]{\includegraphics[width=0.33\textwidth]{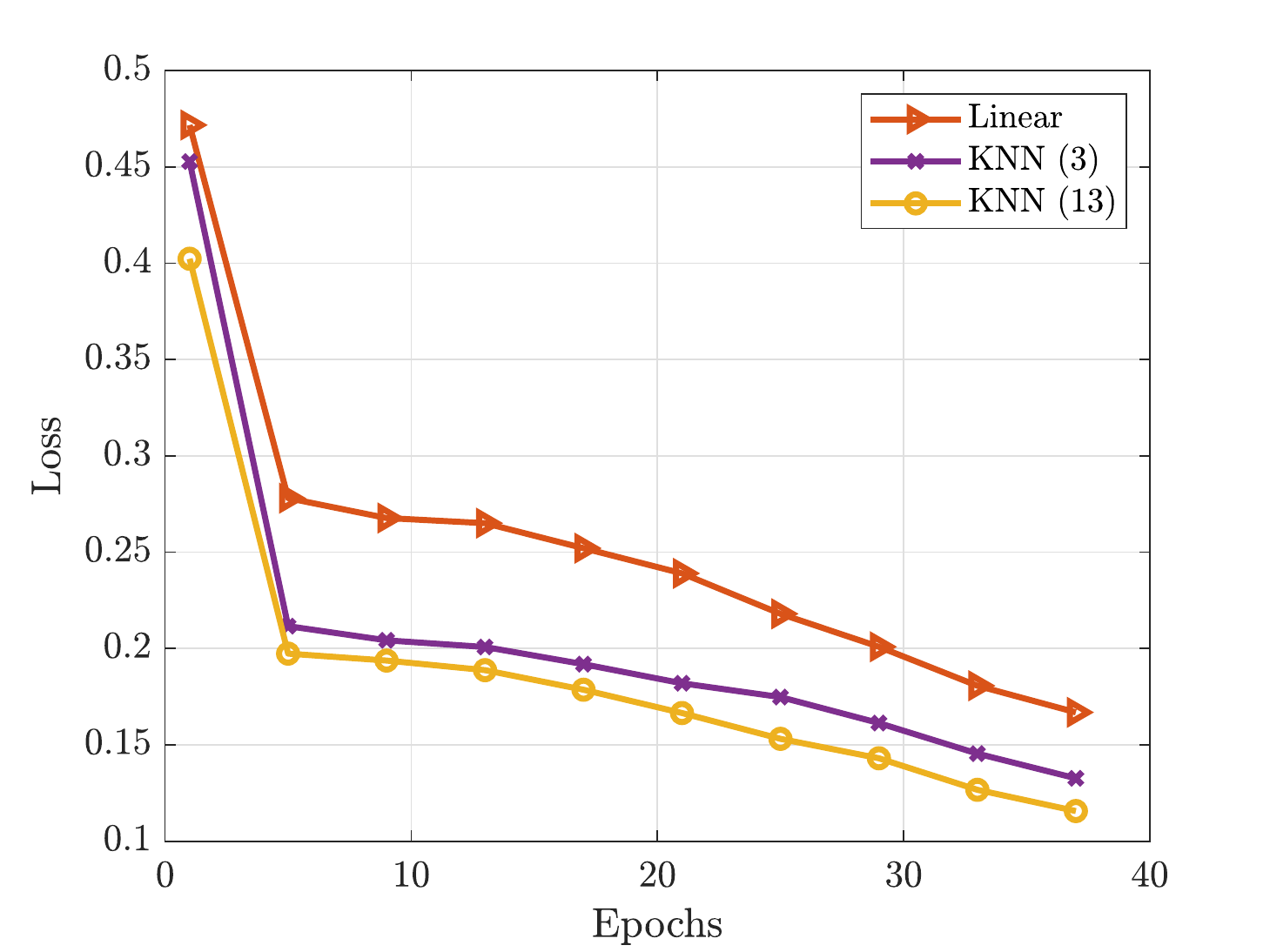}}
    \hskip -1.9ex
    \subfloat[RMSE \label{RMSE}]{\includegraphics[width=0.33\textwidth,trim={0 0 0 0},clip]{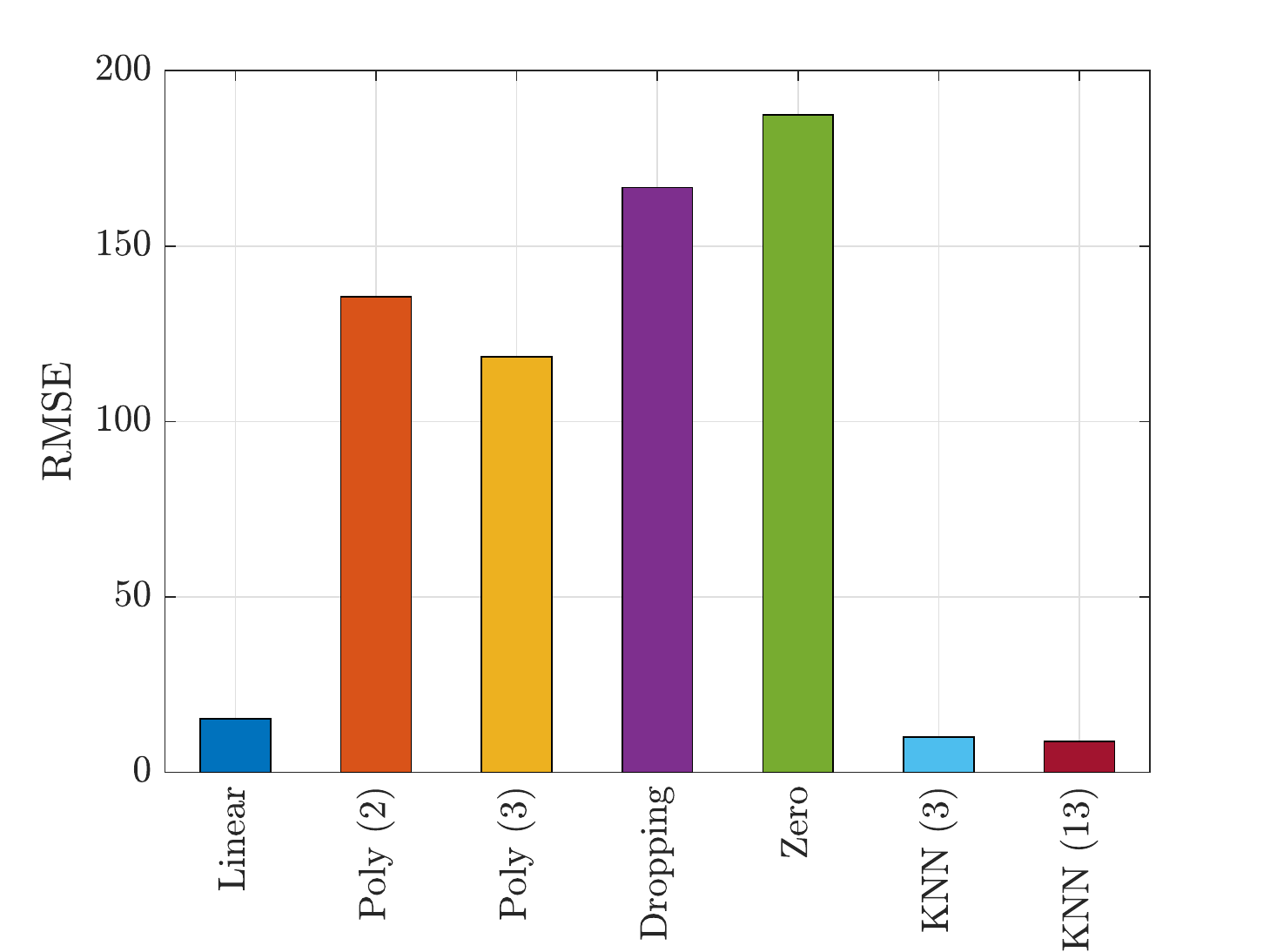}}
    \hskip -1.9ex
    \subfloat[CO2 forecasting \label{Co2_forecasting}]{\includegraphics[width=0.33\textwidth,trim={0 0 0 0},clip]{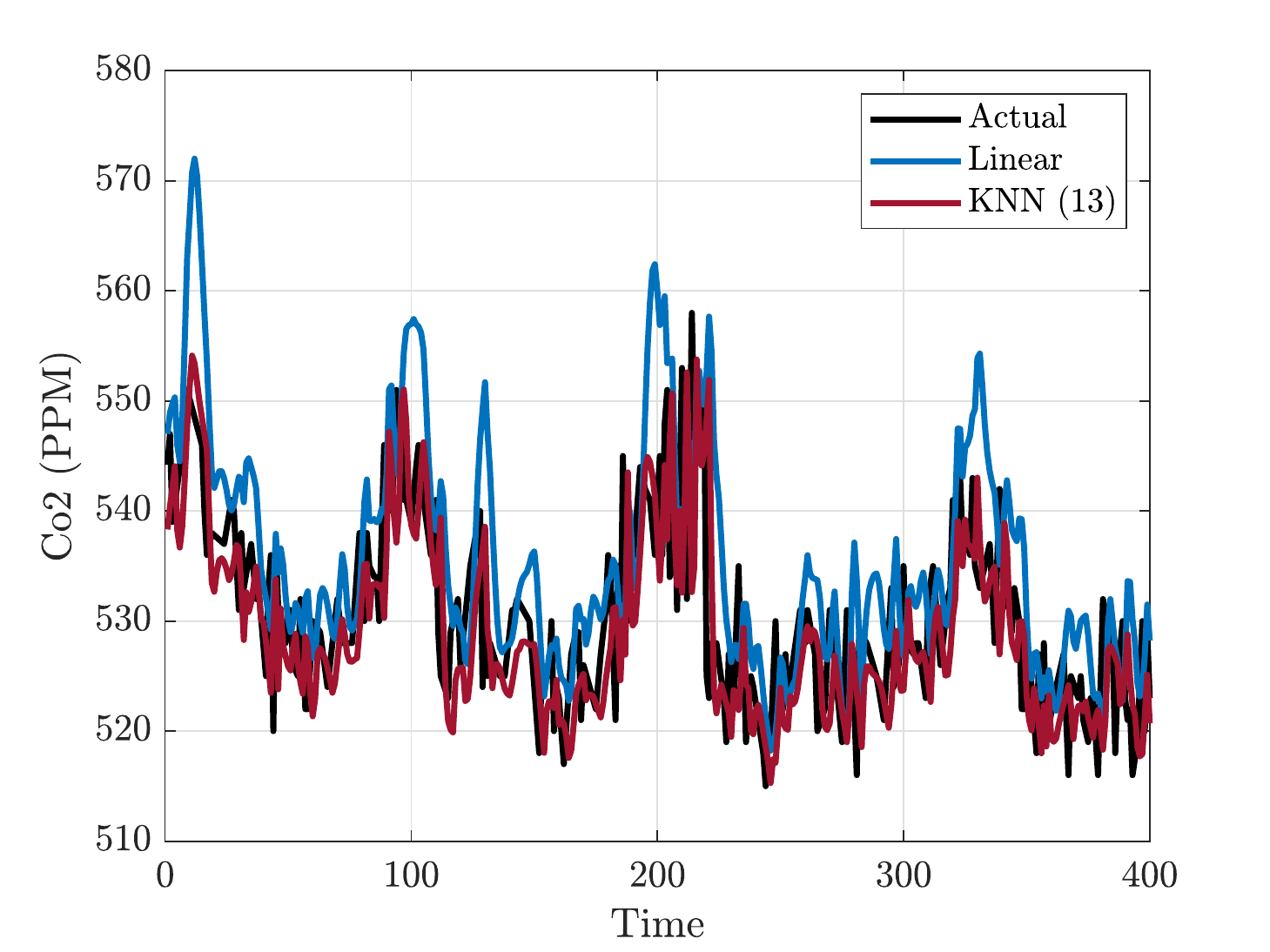}}
    \hskip -1.9ex
    \caption{The LSTM training loss, RMSE, and forecasting future CO2 readings.}
    \label{LSTM_RMSE_FORE} \vspace{-0mm}
\end{figure*}

\section{Performance Evaluation Metrics}\label{KPIs}
In this section, we present the key performance indicators (KPIs):

\begin{enumerate}
    \item LSTM training loss: the LSTM tries to link the relationship between the past and the future through the model weights of the gates. The training loss measures how well the optimized weights describe this relationship on the known test data. The most popular loss function in time-series problems is the mean-square error (MSE)~\cite{7860338}:
    \begin{equation}
    MSE = \frac{1}{M} \sum_{m=1}^D(Y(m)-\tilde{Y}(m))^2,
    \end{equation}
    where $M$ is the size of the data, $Y$ is the true output, and $\tilde{Y}$ is the output of the model.

    \item Root-mean-square error (RMSE): we use the RMSE to evaluate the different techniques for handling the missing data. After handling the missing data, we apply the optimized LSTM to predict a clean (without missing data) future data. Then we calculate the RMSE, which is formulated as:
    \begin{equation}
    MSE = \sqrt{\frac{1}{M} \sum_{m=1}^D(Y(m)-\tilde{Y}(m))^2}.
    \end{equation}
    
    \item Confusion matrix: it depicts each class's true and wrong classifications.

    \item Precision $P$ and recall $R$: the former is the ratio of the correct classified data of a class to the total classified data of that class, whereas the latter presents the ratio between the correct classified data of a class to the total number of data of that class.

    \item f1-score: it is used to evaluate the classification by combining precision and recall. It is calculated as follows:
    \begin{equation}
    \text{f1-score} = \frac{2 \: P  \: R}{P + R}.   
    \end{equation}
    
    \item Total accuracy: the ratio between all the correct classified data of all classes to the total number of data.    
\end{enumerate}

\section{Results and Discussion}\label{results}
This section presents the simulation results of the proposed data analysis techniques. First, we present the failure identification and handling approaches. Then, we test each approach by predicting the readings of the devices in the future. Finally, we use the results to establish a people counter approach. We collect the readings of all the devices within the period $01.02.2020 \: - \: 01.06.2021$. Moreover, we tested the proposed methods on CO2 devices in closed rooms, using the Pytorch framework on a single NVIDIA Tesla V100 GPU and 10 GB of RAM. Linear stands for linear interpolation, Poly $(2)$ is the polynomial interpolation of order $2$, Poly $(3)$ is the polynomial interpolation of order $3$, Dropping is ignoring the missing transmission, Zero is replacing the missing value with $0$, KNN $(k)$ stands for applying KNN prediction to the missing values with $k$ nearest neighbors.



To choose the optimum value of $k$ in the KNN algorithm, we use a mini-batch with no missing transmissions and test its prediction using different values of $k$, then choose the one with the best prediction, e.g., $k = 13$ in our setup. We apply different methods to handle the missing values on the Co2 readings and use the result to predict known Co2 readings that have no missing values. We use LSTM with four layers, each having $128$, $64$, $64$, and $32$ neurons, respectively. We use the MSE loss function and RMSprop optimizer to update the weights of the LSTM. Fig.~\ref{Models_loss} shows the convergence of training the LSTM using linear interpolation, KNN (3), and KNN (13).



\begin{figure*}[t]
    \centering
    \subfloat[Accuracy \label{Acc}]{\includegraphics[width=0.47\textwidth]{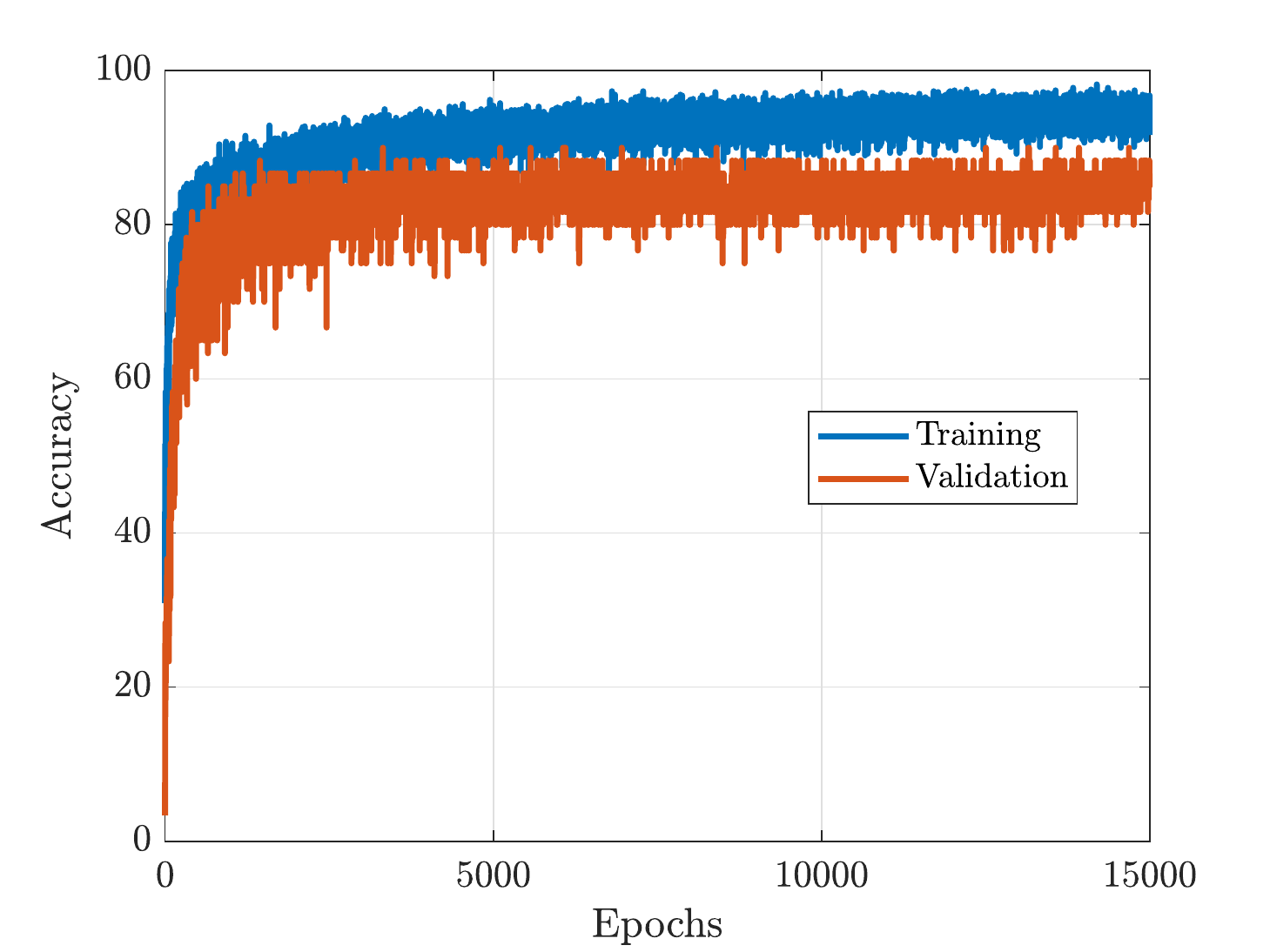}}
    \hskip -1.9ex
    \subfloat[Loss\label{Loss}]{\includegraphics[width=0.47\textwidth,trim={0 0 0 0},clip]{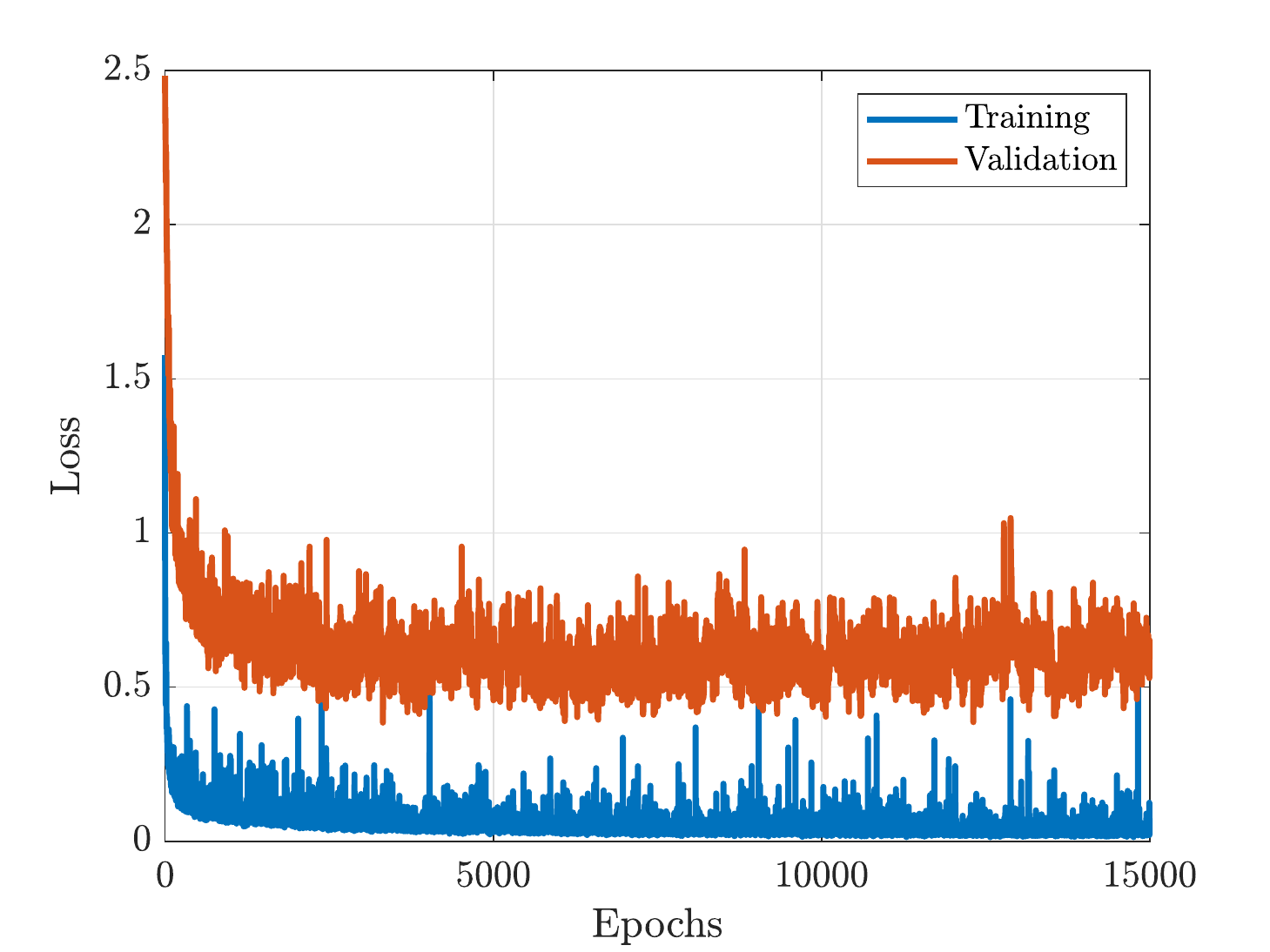}}
    \hskip -1.9ex
    \caption{The training and validation accuracy and loss.}
    \label{Acc_Loss} \vspace{-0mm}
\end{figure*}

Fig.~\ref{RMSE} shows the RMSE between the predicted readings of the LSTM and the actual readings of the Co2 device using different methods to handle the missing values. Using KNN $(13)$ has the lowest RMSE compared to other methods as KNN $(13)$ is considered the optimum approach. Both linear interpolation and KNN $(3)$ has better RMSE results than Poly $(2)$, Poly $(3)$, Dropping, and Zero methods. Keeping the missing transmission as $0$ seems to confuse the predicting model, as it has the worst RMSE values among all the other methods. In Fig.~\ref{Loss}, we present the training loss of LSTM in predicting the future Co2 readings using linear interpolation, KNN $(2)$, and KNN $(3)$ to handle the missing values. We can notice that the KNN $(13)$ has the lowest training loss compared to KNN $(3)$ and linear interpolation, which interprets the high efficiency of using the optimized K in KNN predictors to handle missing values as it fills the missing values accurately and hence, predicts the future efficiently. In both cases, Fig.~\ref{Co2_forecasting} compares the KNN $(13)$ and the linear interpolation in handling the missing values by plotting the predicted Co2 using the LSTM predictor. We notice that the KNN $(13)$ captures the behavior of the true Co2 readings efficiently, outperforming the linear interpolation.

\begin{figure*}[t]
    \centering
    \subfloat[Number of people before oversampling \label{People_cnt}]{\includegraphics[width=0.47\textwidth]{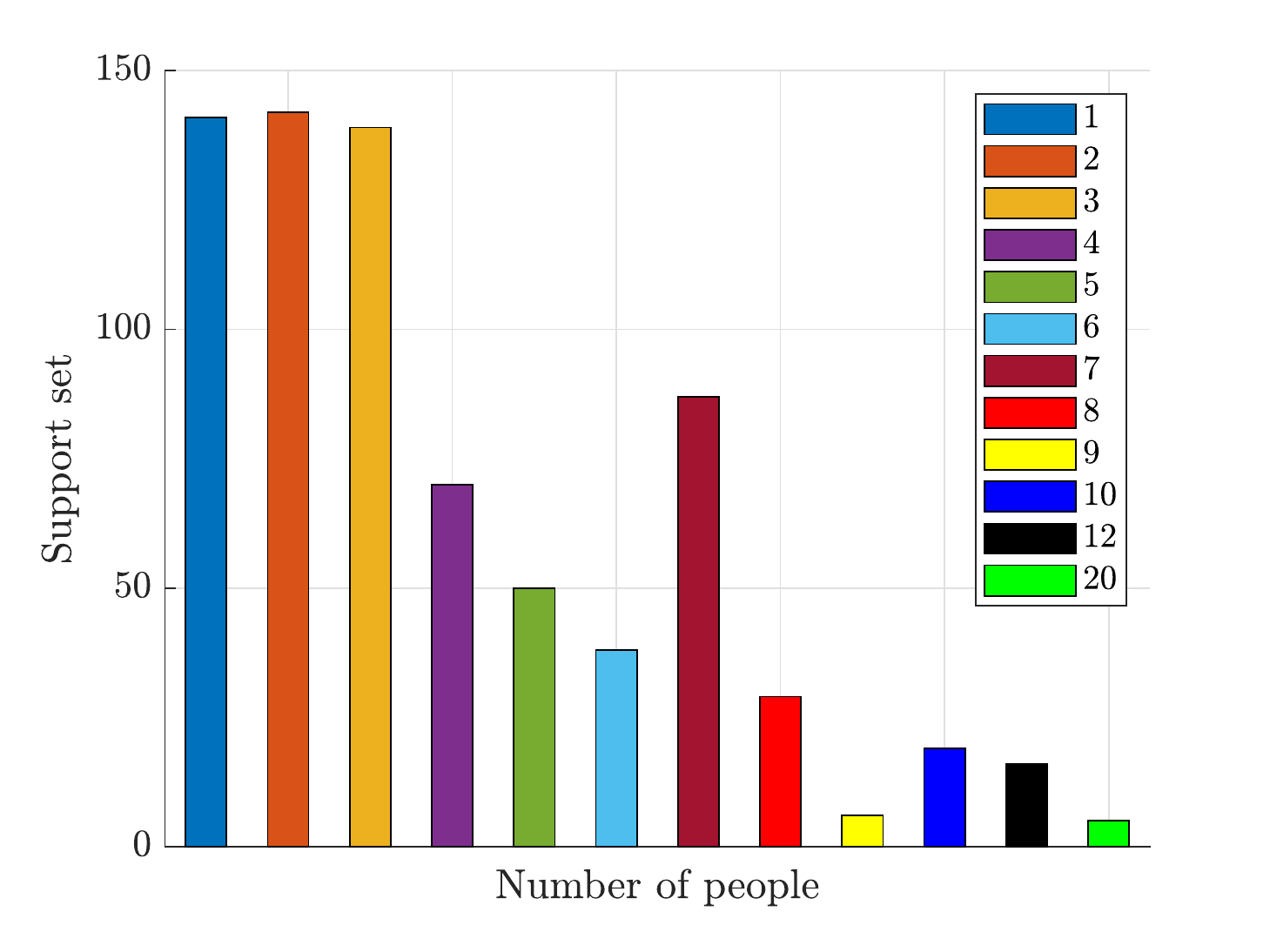}}
    \hskip -1.9ex
    \subfloat[Number of people after oversampling \label{People_cnt_oversample}]{\includegraphics[width=0.47\textwidth,trim={0 0 0 0},clip]{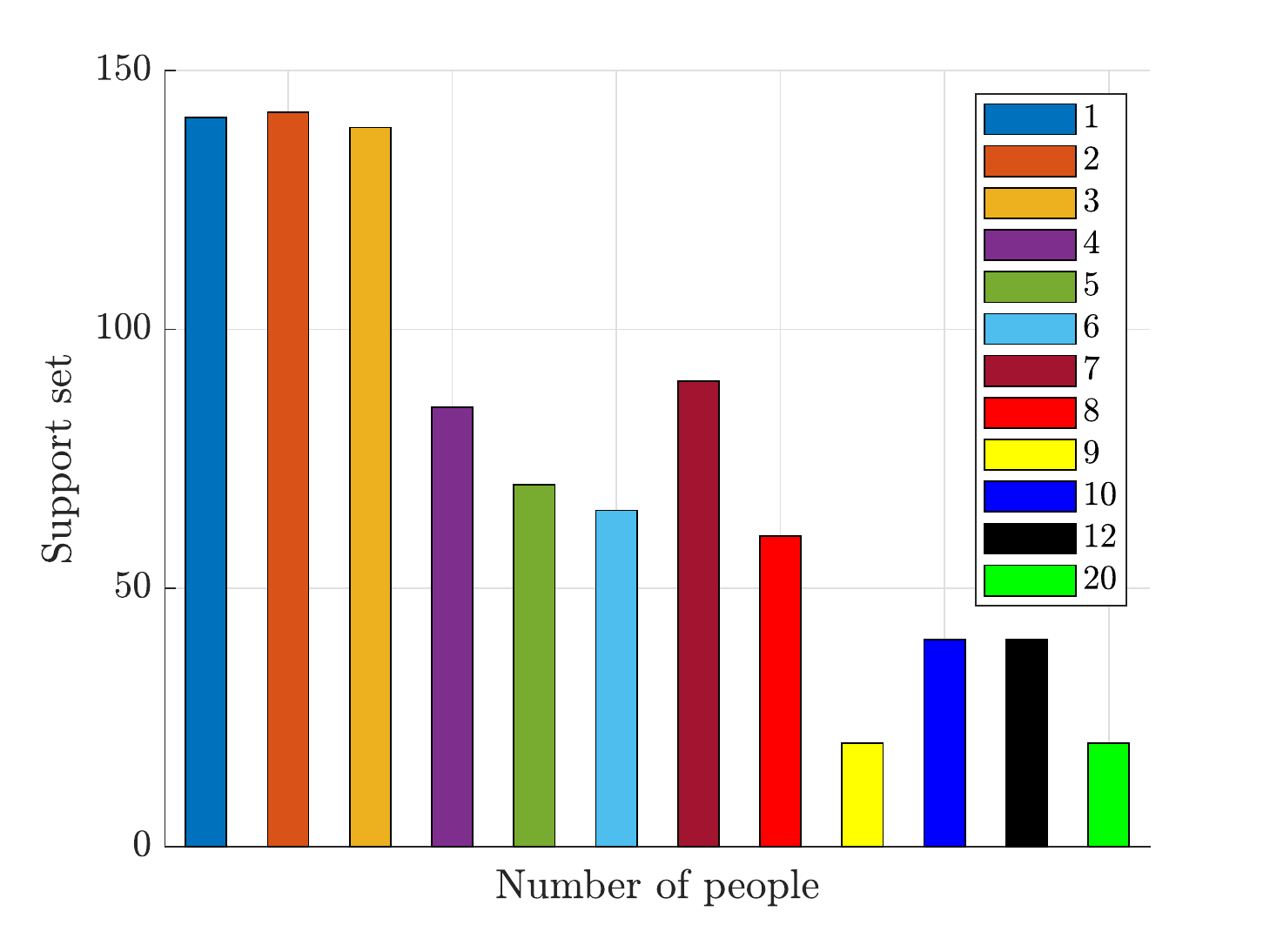}}
    \hskip -1.9ex
    \caption{The support set of all classes.}
    \label{Support_set} \vspace{-0mm}
\end{figure*}

Fig.~\ref{People_cnt} presents the support set collected in a room while recording the device readings found in this particular room. As illustrated before, the features are the readings of the devices, whereas the labels (classes) are the number of people inside the room. We can notice that the number of records for each class is imbalanced. Therefore, we do a pre-processing step, where we oversample the minority classes by duplicating random shots from the minority classes. As a result, we approach almost a uniform distribution of the support set of the classes as shown in Fig~\ref{People_cnt_oversample}. Afterward, a normalization step is done over the dataset and splitting the data into training, validation, and testing datasets. In addition, we ensure that the duplicated data created in the oversampling step exists only in the training set. We build a deep neural network with an input layer with a size equal to the readings of the Co2 sensor, four hidden layers each has $512$, $256$, $128$, and $64$ neurons, and an output layer with a size equals to the number of classes $(12)$ and has a softmax function. We use the cross-entropy loss function and Adam optimizer to update the weights of the neural networks. In Fig.~\ref{Acc}, we present the training and validation accuracy, whereas in Fig.~\ref{Loss}, we present the training and validation loss of the trained neural network. As shown, the model converges (the accuracy increases and the loss decreases) as we train more epochs.

\begin{figure}[t!]
    \centering    \includegraphics[width=1\columnwidth]{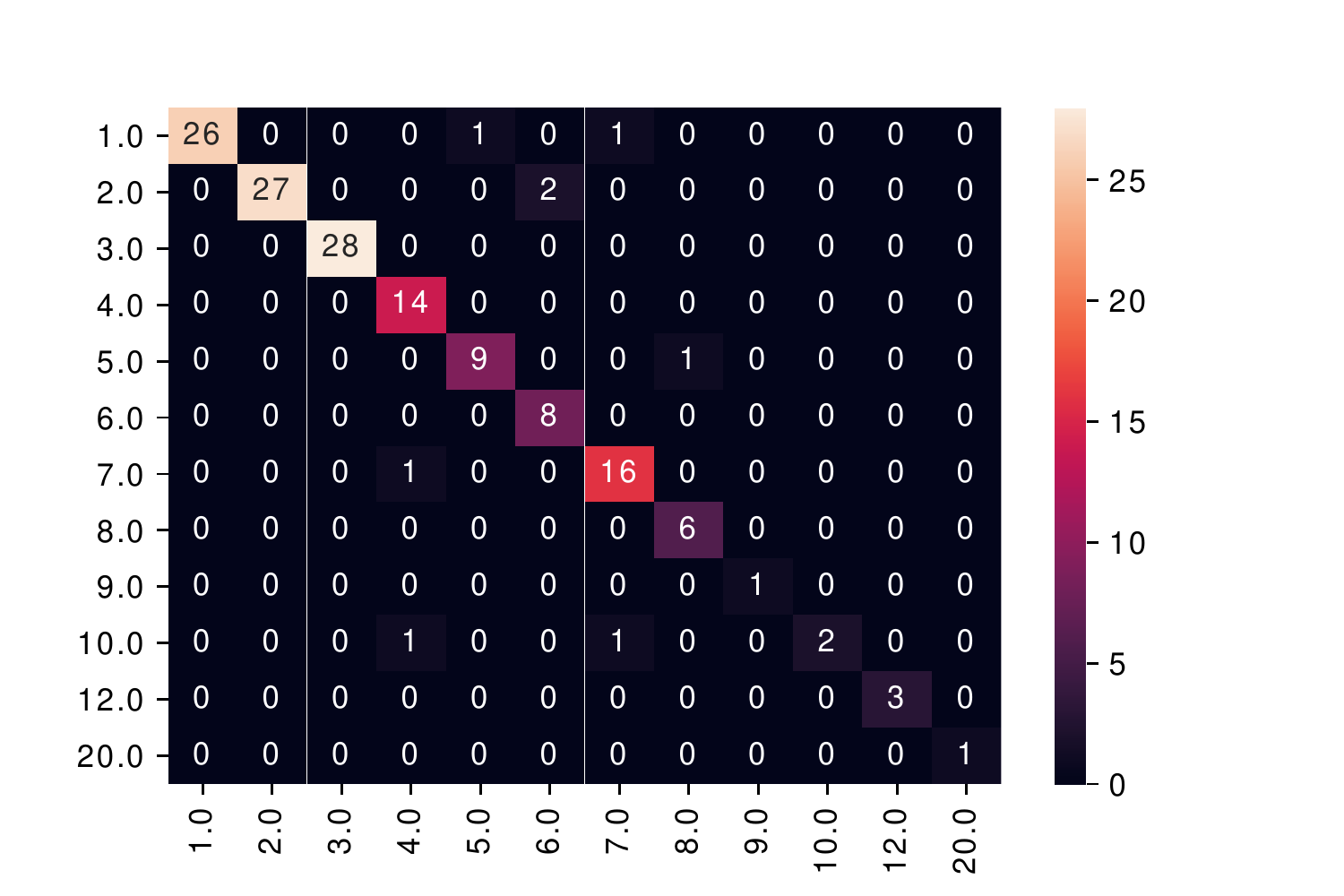} \vspace{1mm}
	\caption{The resulting classification heatmap.} \vspace{0mm}
	\label{Classification}
\end{figure}

Fig.~\ref{Classification} depicts the heatmap (confusion matrix) of testing the trained neural network on the testing set. The right-angled diagonal has the highest numbers in each row, which means the model successfully classifies the test data. Furthermore, Table~\ref{Classification_Report} presents the classification report of the trained neural network, where we show the precision, recall, f1-score of each class, and the overall accuracy. We can notice that apart from class $10$, all the model has good precision, recall, and f1-score for all the classes. The overall accuracy is $95 \%$.

\begin{table}[!t]
\centering
    \caption{The classification report of testing the trained neural network.}
	\label{Classification_Report}
\begin{tabular}{|l|l|l|l|l|}
\Xhline{2\arrayrulewidth}
Class & Precision & Recall & f1-Sc. & Support\\ 
\Xhline{2\arrayrulewidth}
$1$ & $1.00$ & $0.93$ & $0.95$ & $28$\\
\hline                              

$2$ & $1.00$ & $0.93$ & $0.96$ & $29$\\
\hline                              

$3$ & $1.00$ & $1.00$ & $1.00$ & $28$\\
\hline                              

$4$ & $0.88$ & $1.00$ & $0.93$ & $14$\\
\hline                              

$5$ & $0.90$ & $0.90$ & $0.90$ & $10$\\
\hline                              

$6$ & $0.80$ & $1.00$ & $0.89$ & $8$\\
\hline                              

$7$ & $0.89$ & $0.94$ & $0.91$ & $17$\\
\hline                              

$8$ & $0.86$ & $1.00$ & $0.92$ & $6$\\
\hline                              

$9$ & $1.00$ & $1.00$ & $1.00$ & $1$\\
\hline                       

$10$ & $1.00$ & $0.50$ & $0.67$ & $4$\\
\hline                              

$12$ & $1.00$ & $1.00$ & $1.00$ & $3$\\
\hline                              

$20$ & $1.00$ & $1.00$ & $1.00$ & $1$\\

\Xhline{2\arrayrulewidth}
 & & & Acc & $0.95$\\
\Xhline{2\arrayrulewidth}
\end{tabular} \vspace{-0mm}
\end{table}

\section{Conclusions}\label{conclusions} 
This paper presents an analysis of the Smart Campus dataset based on LoRaWAN. First, we identify missing values due to failures in transmission. Then, we perform a comparison between different techniques to fill in the missing values, where the linear interpolation and the KNN show the best results in terms of future prediction on clean data. Furthermore, we build a neural network to predict the number of people inside a room based on the readings of the sensor. The numerical results show the high accuracy of predicting the number of people with $95 \: \%$ accuracy.

We note that the dataset is quite rich, with many features yet unexplored. We hope this work serves as an overview and entry point for further exploration and exploitation of the dataset for many other use cases.  

%

\bibliographystyle{IEEEtran}
\bibliography{IEEEabrv,di}
\end{document}